\documentclass[sn-mathphys-num, iicol]{sn-jnl}

\usepackage{graphicx}%
\usepackage{multirow}%
\usepackage{amsmath,amssymb,amsfonts}%
\usepackage{amsthm}%
\usepackage{mathrsfs}%
\usepackage[title]{appendix}%
\usepackage{xcolor}%
\usepackage{textcomp}%
\usepackage{manyfoot}%
\usepackage{booktabs}%
\usepackage{algorithm}%
\usepackage{algorithmicx}%
\usepackage{algpseudocode}%
\usepackage{listings}%

\usepackage{amsmath}
\usepackage{gensymb}
\usepackage{subcaption}
\usepackage{tabularx}
\usepackage{multirow}
\usepackage{tablefootnote}

\usepackage{siunitx}
\usepackage{bbm}
\usepackage[noabbrev]{cleveref}
\graphicspath{{fig/}}

\newcommand{\comment}[1]{}
\newcommand{\parag}[1]{\vspace{1mm} \noindent {\bf #1}}
\newcommand{\Loss}{\mathcal{L}}
\newcommand{\LDat}{\Loss_{\rm dat}}

\newcommand{\LMalis}{\Loss_{\texttt{MALIS}}}

\newcommand{\Feat}{F}
\newcommand{\x}{\mathbf{x}}

\newcommand{\II}{\mathbbm{1}}

\newcommand{\bI}{{\bf I}}

\newcommand{\ConvNextTiny}{{ConvNeXt\mbox{-}Tiny}}
\newcommand{\SmallDDLN}{{DDLN\mbox{-}S}}

\newcommand{\MediumDDLN}{{DDLN\mbox{-}M}}
\newcommand{\LargeDDLN}{{DDLN\mbox{-}L}}

\newcommand{\HugeDDLN}{{DDLN\mbox{-}H}}

\newcommand{\TheirModel}{{DeepWireCNN}}
\newcommand{\TheirModelAbbr}{{DWCNN}}
\newcommand{\RTDETR}{{RT-DETR}}

\newcommand{\ClampedReLU}{{clamped\mbox{-}ReLU}}
\newcommand{\TheirDataset}{{NE-VBW}}
\newcommand{\OurDataset}{{DDLN}}
\newcommand{\FOne}{{F\textsubscript{1}}}

\begin{document}
	
	\title[Vision-Based Power Line Cables and Pylons Detection for~Low Flying Aircraft]{
		Vision-Based Power Line Cables and Pylons Detection for~Low Flying Aircraft
	}

	\author[1]{\fnm{Jakub} \sur{Gwizda\l{}a}}
	\author[1]{\fnm{Doruk} \sur{Oner}}
	\author[1]{\fnm{Soumava} \sur{Kumar Roy}}
	\author[1]{\fnm{Mian Akbar} \sur{Shah}}
	
	\author[2]{\fnm{Ad} \sur{Eberhard}}
	\author[2]{\fnm{Ivan} \sur{Egorov}}
	\author[2]{\fnm{Philipp} \sur{Kr{\"u}si}}
	\author[2]{\fnm{Grigory} \sur{Yakushev}}\email{gy@daedalean.ai}

	\author*[1]{\fnm{Pascal} \sur{Fua}}\email{pascal.fua@epfl.ch}
	
	\affil*[1]{\orgdiv{Computer Vision Laboratory}, \orgname{EPFL}, \orgaddress{\country{Switzerland}}}
	\affil[2]{\orgname{Daedalean AG}, \orgaddress{\city{Z{\"u}rich}, \country{Switzerland}}}
	

\abstract{

Power lines are dangerous for low-flying aircrafts, especially in low-visibility conditions. Thus, a vision-based system able to analyze the aircraft's surroundings and to provide the pilots with a ``second pair of eyes'' can contribute to enhancing their safety. To this end, we develop a deep learning approach to jointly detect power line cables and pylons from images captured at distances of several hundred meters by aircraft-mounted cameras. In doing so, we combine a modern convolutional architecture with transfer learning and a loss function adapted to curvilinear structure delineation. We use a single network for both detection tasks and demonstrate its performance on two benchmarking datasets. We have also integrated it within an onboard system and run it in flight. We show with our experiments that it outperforms the prior distant cable detection method~\cite{Stambler19} on both datasets, while also successfully detecting pylons, given their annotations are available for the data.
}

	
	\maketitle

\section{Introduction}
\label{sec:introduction}

Low-flying aircraft, such as light planes and helicopters, are always at risk of colliding with ground obstacles. Power lines are especially dangerous because they are hard to see, even in good weather. In 2010, the Federal Aviation Administration used 13 years worth of accident data from the National Transportation Safety Board (NTSB) database to estimate that there are 76 collisions with power line per year on average and that 30\% of them resulted in fatalities~\cite{Miller10}. Utility poles and transmission towers, which we will collectively refer to as {\it pylons}, constitute additional dangers and are often found in the vicinity of power line cables. Many of the accidents stem from the fact that pilots are expected to simultaneously carry out many tasks---steering the aircraft, navigating, monitoring the weather, and performing mission-related tasks---that all require a high level of concentration. This sometimes causes cognitive overload and loss of situational awareness~\cite{Flanigen23}. An automated system that would alert the pilots to an impending collision danger would therefore be extremely valuable. 

Systems that rely on a database of known obstacles, such as FLARM~\cite{FLARM}, are a good start. However, there always is the possibility of obstacles not being in the database. Thus, there has long been interest in developing video-based algorithms to detect power line cables. Most early work relied on image processing methods designed for line detection, both for detection of cables~\cite{Campoy01, Rangachar02, Yan07, Candamo09} and pylons~\cite{Tilawat10}. Feature-based machine learning techniques then started being used~\cite{Rangachar02, Sampedro14, Ceron17} but were superseded by deep-learning based methods, both for power line cables~\cite{Pan16, Gubbi17, Lee17c, Li18l, Madaan17} and transmission towers~\cite{Bian18a}. Most of prior research focuses on either the detection of power line cables or that of pylons~\cite{Zhu23} and does not treat them jointly. Methods for power line cables detection are frequently developed in the context of power line infrastructure inspection, where the cables are likely to be more clearly visible than at larger distances encountered during regular small aircraft flights. In this work we tackle this latter challenging scenario with the aim of developing an aircraft safety system assisting the pilots in detecting these notorious dangers. 

We propose a new deep-learning based approach to jointly detect power line cables and pylons. We use a pre-trained ConvNeXt~\cite{Liu22e} as our backbone and formulate the detection task as one of regressing a map of distances to the obstacles from the images. To validate our method we compare it with the approach proposed by \citet{Stambler19} and use the accompanying \TheirDataset{} dataset -- a public high-resolution dataset created for vision-based power line cables detection in the context of developing aircraft safety system. We demonstrate that combining transfer learning, a modern backbone, and a suitable loss function designed for curvilinear structure detection boosts performance beyond the previously proposed method~\cite{Stambler19} on their dataset. 

There are other datasets that could be used to evaluate cable detection algorithms~\cite{Candamo09, Yetgin18a, Yetgin18b, Yetgin19, Zhang19h, Abdelfattah21, Choi22, Chiu23}. However, they have various limitations, including: a viewing perspective from the level close to the ground and mostly in urban settings \cite{Candamo09, Chiu23}, scaled down higher resolution images with only image-level labels \cite{Yetgin18a, Yetgin18b}, small dataset size \cite{Choi22} or well visible power lines \cite{Zhang19h, Yetgin19}, including scenarios intended for testing cable inspection methods with power lines seen from a short distance \cite{Abdelfattah21}. We are facing a very different scenario with broad fields of view, distant barely visible power lines and high-resolution images. To develop our detection method we created our more diverse {\OurDataset} dataset, recorded with short focals that provide information about a large area in front of the aircraft. We also evaluate our approach on it and demonstrate the strong performance of the proposed method. Finally, we show that, given additional annotations for pylons, a single network can easily and successfully be trained to detect them in addition to cables. Furthermore, when only pylon detection is needed and small detection errors are tolerated, it performs competitively to the state-of-the-art real-time object detector~\cite{Zhao24} of similar size.

In the following \cref{sec:related} we present prior works on vision-based power line cables and pylons detection. In \cref{sec:method} we describe our model architecture and the loss function. \Cref{sec:results} presents our results, accompanied by the descriptions of the datasets used for the evaluation, baselines we compare with and the implementation details. Finally, we summarize our work and discuss the limitations of the presented method together with the possible future directions in \cref{sec:conclusion}.


\section{Related works}
\label{sec:related}

\subsection{Detecting Power Lines}

Vision-based detection of power line cables has a long history. Many approaches, including some relatively recent ones, rely on image processing techniques, such as edge detection or the Hough and Radon transforms~\cite{Campoy01, Rangachar02, Yan07, Candamo09, Song14, Tian15, Baker16, Santos17, Oh17, Ganovelli18, Zhou16}. 

Such detection methods that rely on manually designed image processing pipelines have now been outperformed across the whole computer vision field by learning-based methods, provided that sufficient amount of training data is available. Thus, the approach to cable detection has changed with the advent of deep learning techniques. Several works formulated the detection task as one of classifying image patches according to the presence or absence of power line cables. In~\cite{Pan16}, previously extracted image edge maps are used as the input to the trained model. In~\cite{Gubbi17}, Histogram of Gradient (HoG) features are used instead. The algorithm of~\cite{Lee17c} performs patch classification and recovers pixel-wise segmentation indirectly by processing intermediate network activations. That of~\cite{Li18l} follows a hybrid approach with a hierarchical CNN classifier over a set of different resolution of patches, followed by traditional techniques based on edge filters and a Hough Transform variant to precisely localize wires. These classification approaches were by design weakly-supervised~\cite{Lee17c} for cables presence and could not benefit from a richer information about detected objects' appearance, which can be provided by binary masks and exploited by segmentation-based approaches.

Another line of research focused on using object, line or key points detectors. \citet{Nguyen20} proposed LS-Net, a single-shot detector, combining classification and line endpoints regression tasks, performed over a set of four overlapping grids of cells. Their method has been evaluated on a dataset~\cite{Yetgin19} featuring 800 crops extracted from full HD frames, scaled to $512\times512$ px patches, with power line cables clearly visible over the terrain background. \citet{Abdelfattah21} introduced the TTPLA dataset, featuring high-resolution aerial images captured by a drone flying within a close distance from power line infrastructure and annotated with cable and transmission tower masks. They also established an instance segmentation baseline with the use of YOLACT~\cite{Bolya19}, an extension of single-shot object detector with a segmentation capability. Recently, \citet{Xing23} designed a power line detection method for use by inspection drones. They used YOLO~\cite{Redmon16}, trained solely on synthetic data which additionally classified the inclination of a power line cable within the detected bounding box. However, as in~\cite{Abdelfattah21}, their data was restricted to close views of power line cables from the perspective of a drone hovering above the inspected infrastructure. Finally, a key point detection scheme was proposed by \citet{Dai20}. They formulate the detection as a regression of 5 key points evenly distributed across each cable in the input image. The points regressed by the network are subsequently clustered and then a curve representing the complete power line cable can be fitted to their locations. Their method was evaluated on PLDU and PLDM datasets~\cite{Zhang19h}, which contain in total 860 images with $540\times360$ px resolution, each of which with several wires to be detected. These datasets feature again very (PLDU) and relatively (PLDM) close views of cables, often clearly visible against an urban (PLDU) or a mountainous (PLDM) scenery. This makes the cable detection task easier than in our challenging setting of detections at significant distances and over complete high-resolution image.

To leverage the precise supervision signal pixel-wise binary masks can provide, multiple methods formulate power line cable detection as a segmentation task. \citet{Madaan17} applied a CNN directly to the images containing wires and treated detection as a semantic segmentation task. \citet{Stambler19} also performed power line detection relying on segmentation objective, however trained jointly with a regression of parameterized line segment proposals. Both of these methods use custom network architectures, including the use of dilated convolutions and trained from scratch. An alternative approach involves reusing a CNN model pre-trained on a large image datasets and fine-tuning for the cable detection task. This was demonstrated by \citet{Zhang19h} with a trimmed VGG16 model~\cite{Simonyan15, Liu17a} pre-trained on the ImageNet dataset \cite{Deng09}, but only on the comparatively easy PLDU and PLDM datasets~\cite{Zhang19h}. We follow a similar strategy in our own approach.

Other more recent segmentation approaches to power line cables detection included models based on UNet~\cite{Jaffari21} or UNeXt~\cite{Cheng23b} architectures. Others leverage generative adversarial networks (GANs) trained to highlight power lines in the images~\cite{Abdelfattah23} or added transformer blocks to CNN feature extractors enabling global attention~\cite{An23}. Most of these methods \cite{Cheng23b, Abdelfattah23, An23} were evaluated on the TTPLA dataset~\cite{Abdelfattah21}, which features high-resolution close views of cables and pylons, mostly suitable for power line infrastructure inspection. \citet{Jaffari21} evaluated their model on IVRL~\cite{Yetgin19} and PLDU \cite{Zhang19h}, both containing well visible cables crossing the images and the former with low variability of wire appearance. \citet{Cheng23b} additionally measured their performance on the VITL dataset \cite{Choi22} which is rather scarce with only 400 crops of higher-resolution images, with a size of $256 \times 256$ pixels each. In this work, we evaluated the performance of the models on two datasets -- \TheirDataset{} \cite{Stambler19} and \OurDataset{} dataset collected by us -- both of which comprise complete, high-resolution images captured from the flying aircraft within significant distances from the detected cables and pylons, a setting relevant in the development of a system aiming to improve the aircraft safety.

Cable segmentation models have also been used in the field of computational photography, along with a wire segmentation and in-painting model trained to improve the image aesthetics by removing cables from the photographs. \citet{Chiu23} have collected the WireSegHR dataset containing 6000 very high-resolution images with pixel-wise annotations of wire locations. Judging by the released test set of 420 images, the dataset contains various kinds of wires and cables, along with other elongated objects, often with challenging appearance and poor visibility in a diverse set of urban sceneries and in cluttered scenes. The images were captured from the ground level as in a typical photography setting. This makes for a different viewing perspective than the one encountered by aircraft. In the images from our \OurDataset{} dataset and from the \TheirDataset{} dataset \cite{Stambler19}, on which we evaluated our method, the wires are mostly seen by the aircraft-mounted camera during flight.

In our approach to localizing power line cables instead of segmentation we use distance mask regression, which is a known choice for curvilinear structures detection~\cite{Meyer18, Davari22, Xue23}. As such, our method is closest to the regression of line segment proposals in~\cite{Stambler19}. However, we use a modern convolutional network architecture \cite{Liu22e}, leverage transfer learning by starting with a model pre-trained on the ImageNet dataset~\cite{Deng09} and extend the regression loss function with a component promoting connectivity in the detections, introduced in our earlier work~\cite{Oner22a}. 

\subsection{Detecting Pylons}

The detection of transmission towers and utility poles has also been the focus of much previous work. Close-range and mid-range detection is usually required in the contexts of inspection~\cite{Bian18a, Watanabe18, Wang19n, Yang23a}, mapping~\cite{Zhang18f}, damage assessment~\cite{Chen20e, Mo21} and aircraft safety~\cite{Zhang23c}. In contrast, very long-range detection in satellite or aerial images is mostly performed in top-down views~\cite{Gomes20, Haroun21, Wang21h, Zhang22, Zha23}. There has been less interest in frontal detection in aerial views at a distance of hundreds of meters, which is the focus of this paper. 

Like power lines, pylons can be treated either as thin linear objects or as ones comprised of many line segments, which is especially useful when dealing with the skeletal structure of larger towers. Therefore, similar image processing tools have been used in early pylon detection work. \citet{Tilawat10} used an IIR filter followed by a Hough transform to detect linear segments in the images. Based on Hough counts in local windows, the probability of pylon presence can be assessed. Similarly, \citet{Araar15} detect line segments in the skeleton of towers using a Line Segment Detector~(LSD)~\cite{Grompone10} and the detections are filtered based on their color characteristics. Other works rely instead on keypoint detectors. \citet{Zhang14d} used SIFT~\cite{Lowe04} key points descriptors and matched them with a set of pylon template features, while \citet{Ceron17} used FAST~\cite{Rosten06} keypoints and ORB~\cite{Rublee11} descriptors for training a SVM for the classification of image regions. \citet{Sampedro14} obtained the HoG pylon features and used them to train two multi-layer perceptron (MLP) networks. The former performs binary classification of image regions for pylon detection, while the latter was used for classification of pylon structure types.

As with power lines, newer techniques rely on deep learning methods for pylon detection. \citet{Bian18a} designed an unmanned aerial vehicle (UAV) navigation system for power line inspection. It relies on detecting transmission towers, for which they used a customized version of Faster R-CNN~\cite{Ren15}. Other works leverage various versions of the YOLO~\cite{Redmon16} detector~\cite{Wang19n, Chen20e, Mo21, Zhang22, Yang23a, Zhang23c, Chen24a}. Detections from an urban road-level perspective also tend to rely on object detectors -- \citet{Watanabe18} used SSD~\cite{Liu16a} and \citet{Zhang18f} used RetinaNet~\cite{Lin17f}. 

In our work, we depart from these standard approaches and formulate the task as distance mask regression, as when detecting cables. We compare distance mask based pylon detection performance with that of \RTDETR{}~\cite{Zhao24}, one of the state-of-the-art real-time object detectors, enhanced with SAHI~\cite{Akyon22} -- inference and fine-tuning strategies for high-resolution images. We show that when distinguishing individual pylons as separate objects is not required and small detection errors are tolerated, our model using similar numbers of learnable network weights delivers a competitive performance, as measured by the metrics evaluating detection outputs. 

Transmission towers, thanks to their overall greater size and larger thickness of their components, should be easier to detect than power line cables. Their presence is naturally related to that of power line cables. Several works have leveraged this presence as a useful contextual cue to facilitate the subsequent detection of power lines. 
A first subgroup of such techniques relies on hierarchical and algorithmic approaches to infer cable presence given tower detections. \citet{Haroun21} infers power line locations from satellite images by simply connecting pairs of closest detected transmission towers forming a single chain of connections. In a more sophisticated algorithm, after detecting the pylons from the point cloud data, \citet{Flanigen23} used the heights, the distances, and the angles formed by a series of detected towers to infer possible cable connections between the pylons. In another hierarchical manner, \citet{Li18n} used pylon detections to restrict the regions of interest for the subsequent detection of cables. In the absence of pylon detections, either because there are none or because of missed detections, they proposed a second, cable-only detection pipeline. These approaches can be understood as using pylon detection as the way of guessing or restricting the area for cables detection. As observed in~\citet{Li18n}, such methods have only limited applicability when pylons can be detected first~\cite{Haroun21, Flanigen23}. Otherwise, when the support points of the hanging cable are not visible, there remains a need for handling cable-only detection ~\cite{Li18n}. In our work, we show that the joint detections of both cables and pylons is feasible using a single deep learning model, without any specific supervision relating both of the detections. 

Another group of techniques link cable and pylon detections more tightly. Combined evidence for cables and pylons presence was used by \citet{Zhang14d} by formulating a Bayesian framework including terms modeling the spatial correlation between the two object types. This formulation was used to determine whether a line detected by a line detector should be considered as a power line cable or not. The method was further developed in \cite{Shan17}. Its flexibility has been increased by replacing the spatial correlations between pylons and cables by an unconstrained set of auxiliary regions, automatically selected in the vicinity of hypothetical power line detections. The results were evaluated on the BHU dataset~\cite{Zhang14d}, which features relatively low resolution images and power line infrastructure seen from ground level. In our work, we tackle a different setting, with power lines infrastructure captured in high-resolution images but further from the camera and seen from the perspective of an aircraft-mounted camera with cables visible against various backgrounds, mostly originating from the terrain behind them.

Most recently, deep learning methods were tried for joint power line cables and pylon detection~\cite{Feng23, Zhencang23, Zhu23, Ma24}. Two of these methods~\cite{Zhu23, Ma24} are instance segmentation approaches that yield detailed segmentation masks for high resolution inputs. Both of them were evaluated on datasets gathered by UAVs. They comprise close views of the power line infrastructure seen from the top. \citet{Zhu23} evaluated their method additionally on the TTPLA dataset~\cite{Abdelfattah21}, which is mostly suitable for power line inspection related tasks. \citet{Feng23} and \citet{Zhencang23} opted for modified YOLOv3~\cite{Redmon18} object detectors. The former used the detection and clustering of auxiliary targets, such as transmission towers, to determine power distribution corridors area and filter out power line detections outside of this zone, a method reminiscent of the algorithmic approaches discussed above~\cite{Haroun21, Li18n}. The method was evaluated on a dataset of low-voltage power lines in urban setting, in close, inspection views of the power line infrastructure. Finally, \citet{Zhencang23} added to an object detector's architecture a {\it relational suggestion network} with an attention mechanism designed to establish relationships between power lines and pylons. While their dataset features more distant views of power lines than others, the images are captured in relatively low resolution of $640\times480$ pixels, by an infrared camera and from a low altitude of at most \SI{60}{\meter} above ground. These are therefore quite different from the ones we use.


\section{Method}
\label{sec:method}

To jointly detect {\it cables}---power and telephone lines---and {\it pylons}---utility poles and transmission towers---we train a {\it single} CNN based on a modern backbone. Taking our inspiration from the delineation work of~\cite{Oner22a}, we train it to produce distance maps, one from the detected cables and other from the detected pylons. This effectively turns the detection problem into a regression one in which we treat a zero distance as a detection. Our network can be used for cable- or pylon-only detection by simply removing one of its heads. 

For cable detection, we differ from the approach of~\citet{Stambler19} by predicting only the undirected distance to the closest wire, as opposed to distance, direction, and a binary value indicating the presence or absence of a cable. The superior performance of our model, which we demonstrate in the result section, is achieved by combining transfer learning, a modern convolutional backbone, and a loss function that favors cable-like topologies.
 
\subsection{Network Architecture}

\begin{figure*}[ht]
	\centering
	\includegraphics[width=\linewidth]{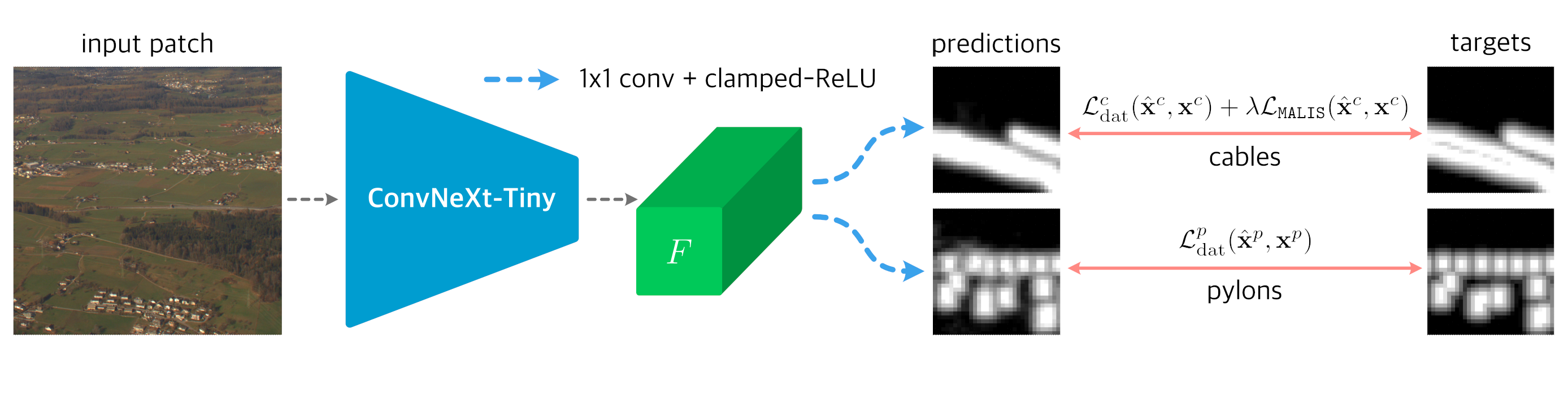}
	\caption{Architecture of the joint power line cables and pylons detection model. Output heads regress cables and pylons distance masks directly from the common feature map $\Feat{}$ generated by the fine-tuned truncated ConvNeXt-Tiny feature extractor.
	\label{fig:architecture}}
\end{figure*}

Fig.~\ref{fig:architecture} depicts the architecture of our model. It comprises two main parts, a pre-trained convolutional feature extractor and two small output heads that regress distances maps from the features.


\parag{Feature Encoder.} 
As our backbone, we use {\ConvNextTiny}~\cite{Liu22e} pre-trained on the {ImageNet}~\cite{Deng09} dataset. We discard its classification head and only use the feature extractor. We can reduce the original model's size by adapting the number of blocks $B$, as defined in~\cite{Liu22e}, in the feature extractor by removing the final ones. In this way, we generated the following variants of our model: 
\begin{itemize}

 \item \SmallDDLN{} (small) with $B = (3, 3, 1)$,
 \item \MediumDDLN{} (medium) with $B = (3, 3, 5)$,
 \item \LargeDDLN{} (large) with $B = (3, 3, 9)$,
 \item \HugeDDLN{} (huge) with $B = (3, 3, 9, 1)$.

\end{itemize}
We do not use the layer normalization and strided convolution following the third stage in the complete {\ConvNextTiny} model. Therefore our models output a $384\times(H/16)\times(W/16)$ feature map $F$, where $(H, W)$ is the resolution of the input image. For \HugeDDLN{} the feature map is $768\times(H/32)\times(W/32)$ and the model requires an additional spatial downsampling level.

\parag{Output Heads.}
They are implemented using pixel-wise regression layers coded as $1\times1$ convolutions. They are applied directly to the feature map $F$ computed by the feature extractor. Distance predictions are obtained using a {\ClampedReLU} activation, restricting the outputs to the $[0, 1]$ range. Zero denotes the detected object, either a cable or a pylon depending on the model head. One corresponds to a set maximal distance value $d_{\rm max}$ that denotes an area sufficiently far away from the objects of interest. In our experiments, we take $d_{\rm max}$ to be 128 pixels at the original input image resolution. 

\subsection{Loss Function} 

Given an image $\bI$, our network outputs two distance maps $\hat{\x}^c$ and $\hat{\x}^p$, one from the cables and the other from the pylons. Let us assume we are given ground-truth versions $\x^c$ and $\x^p$ of these distance maps thresholded at the $d_{\rm max}$ distance introduced above and linearly scaled to the range $[0, 1]$. Our goal is to train the network so that predictions and ground-truth are as similar as possible to each other, while enforcing good connectivity in the cable reconstruction. To this end, we define a composite loss function that is a weighted sum of a data term and a connectivity term, which we describe below. 

\parag{Data Loss Term $\LDat{}$.}
The simplest way to write a data loss term would be in terms of the Mean Square Difference (MSD) between $\hat{\x}^c$ and $\x^c$ on one hand, and $\hat{\x}^p$ and $\x^p$ on the other. However, this would not account for the imbalance between the number of pixels that belong to target structures and the much larger one of those that do not. To do so, let us consider the frequency function $f^k$ that associates to each pixel $\x^k_i$ its distance value occurrence frequency in $\x^k$, where $k$ can be either $c$ for cable of $p$ for pylon. In other words, we write
\begin{equation}
	f^k(\x^k_i) = \frac{1}{N}\sum_{1 \leq j \leq N} \II{}(\x^k_j = \x^k_i) \; ,
\end{equation}
 where $N$ is the total number of pixels in $\x^k$ and the $\x^k$ distance masks are discretized to integer values. We use $f$ to write the weighted version of MSD
\begin{align}
\LDat(\hat{\x}^k,\x^k) &= \sum_{1 \leq i \leq N}  w( \x^k_i)  \| \hat{\x}^k_i- \x^k_i\|^2 \; , \label{eq:LDat} \\ 
w(\x^k_i) &= \left( \log (1 + \epsilon + f^k(\x^k_i)) \right)^{-1} \; ,  \nonumber
\end{align}
 where $\epsilon$, which we set to 0.02, has the purpose of preventing division by values close to 0 and therefore caps the weights at values lower than $\left( \log (1 + \epsilon) \right)^{-1}$. This scheme gives more weight to the sparse features of interest (cables and pylons) and less to the more extensive background regions, mitigating the imbalance in their occurrence frequency. At the same time, it gives greater weights to areas in the $(0, 1)$ range than to the 0-valued areas at the precise location of the detected objects. However, thanks to the logarithmic scaling, the difference in weights remains smaller by an order of magnitude than what it would be using inverse frequency weights and avoids excessive penalization of the most frequent distance values -- 0 and 1, corresponding to target objects and distant background, respectively.

\begin{figure*}[ht]
	\centering
	\includegraphics[width=\linewidth]{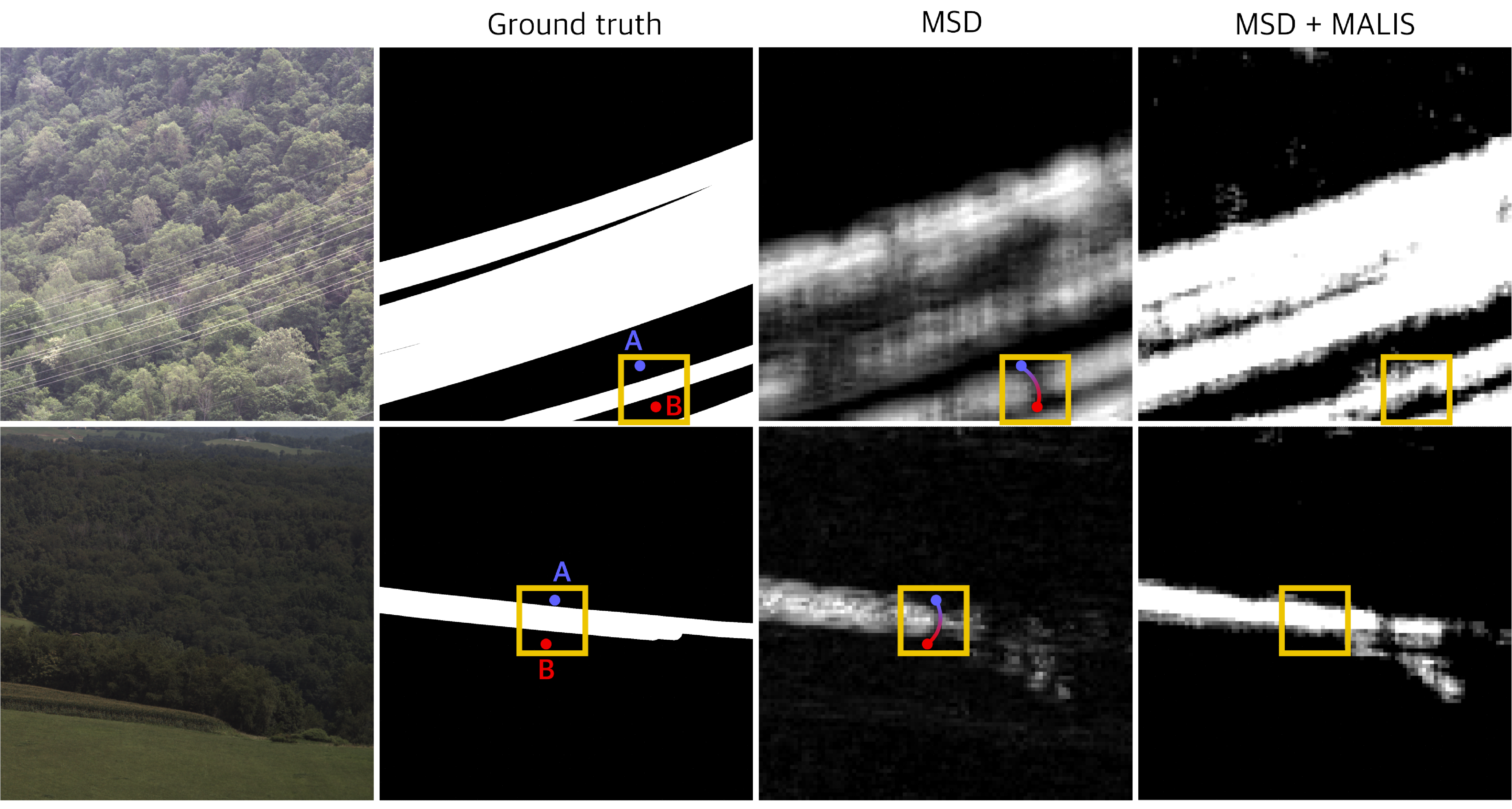}
	\caption{A selection of two outputs for two loss function variants illustrating the application of $\LMalis$ connectivity loss term. The rows contain from left to right: an input image patch, a ground-truth cables segmentation mask, prediction of a model trained with MSD loss alone and prediction of a model trained with MSD loss combined with $\LMalis$ connectivity term. The two last images in each row contain distance masks in which $[0, 1]$ values range was mapped to a grayscale range from white to black. $\LMalis$ is applied to distance mask predictions within the marked square windows. Points A and B belong to two disjoint background regions, which should be separated by the cable detection in between. In the exemplary distance masks predicted for the cables with independent MSD loss supervision these points can be joined with a path traversing through not sufficiently small distance values. This can result in a broken cable detection in the final segmentation, i.e. in a thresholded distance mask. Such segmentation means having an unwanted merger of the background regions which $\LMalis$ aims to prevent. \label{fig:malis}}
\end{figure*}

\parag{Connectivity Loss Terms $\LMalis$.}
The loss function $\LDat{}$ as described above is a pixel-wise loss. When the annotations do not perfectly coincide with the imaged structures, which is often the case, networks trained using such per-pixel losses produce binary masks plagued by topological errors, such as interruptions and missed junctions. In earlier work~\cite{Oner22a}, we introduced a loss function $\LMalis{}$  whose minimization forces to produce distance maps with the same topology as that of the ground truth, even if the actual location of the detection is somewhat imprecise. The difficulty in writing such a loss is to express this requirement in the form of a differentiable loss function that can be used to train a deep network. The central idea of this earlier approach is to forgo enforcing connectivity of the pixels annotated as belonging to linear structures and whose location may be inaccurate. Instead, we express the connectivity of the annotated structures in terms of the disconnections that they create between regions annotated as background. More precisely, we require that two regions separated by a line in the ground truth, are also separated in the prediction. This effectively enforces continuity of the predicted cables. By requiring that connected components of pixels annotated as background remain connected in the prediction, we prevent predicting false positives. To capture dead-ending segments, we compute our loss in small image windows, which are likely to be subdivided even by short cable sections. In other words, we re-purpose the differentiable machinery proposed in the MALIS segmentation algorithm~\cite{Turaga09, Funke18a} to enforce the dis-connectivity of image regions separated by cables, which we illustrate in the \cref{fig:malis}. For more details, we refer the interested reader to our earlier publication. 

\parag{Composite Loss Term $\LDat{}$.}
Given the $\LDat{}$ and $\LMalis{}$ loss terms defined above, we optimize the composite loss function
\begin{align}
\label{eq:cablesAndPolesLoss}
\Loss((\hat{\x}^{c}, \hat{\x}^{p}), (\x^{c}, \x^{p})) &= \LDat^c(\hat{\x}^{c}, \x^{c}) +  \\
\LDat^p(\hat{\x}^{p}, & \x^{p}) + \lambda \LMalis{}(\hat{\x}^{c}, \x^{c}) \nonumber
\end{align}
where upper indices $c$ and $p$ differentiate between cables and pylons and $\lambda$ is a weighting factor for the MALIS loss component, which we set to 0.2.

\subsection{System Deployment}
\label{ssec:hardware}

We deployed DDLN-S, our smallest model variant for joint cables and pylons detection, to Eval Kit, Daedalean's hardware flight assistant system and set it up on a Robinson R44 Raven II helicopter. In order to achieve that, the PyTorch representation of the model and its trained weights were first converted into the ONNX format and then optimized using TensorRT API for the target system and an NVidia GeForce 3060 GPU card. To increase the inference speed, we additionally used half-precision floating point format for the network.

We also developed a pre- and post-processing pipelines, applied to the model's inputs and to its predictions, respectively. The $4096\times3000$ px raw images captured by the camera are debayered, padded and split into $1024\times1024$ px non-overlapping patches. The patches are processed by the GPU in batches of size 4 and then stitched into a $128\times93$ px output representing detections over the complete camera frame. In parallel, dense inverse search optical flow~\cite{Kroeger16, Bradski00} is applied to the two subsequent downsampled camera frames to estimate the movement in the image. Once both results are available, the previous model prediction for the complete frame is warped according to the optical flow output and then it is combined with the prediction for the current frame by computing the average of the two. This averaging operation aims to make the result less susceptible to both false positive and false negative noise. In the following step, we threshold the averaged distance masks turning cable and pylon detections into segmentation masks. Finally, these binary masks are upscaled to the original image size, overlapped with the input image and the result is forwarded to the system's user interface.


\section{Results}
\label{sec:results}

\subsection{Datasets}

There are many publicly available annotated datasets that can be used to validate power line detection algorithms. They include USF \cite{Candamo09}, IVRL \cite{Yetgin18a, Yetgin18b, Yetgin19}, PLDU and PLDM \cite{Zhang19h}, TTPLA \cite{Abdelfattah21}, VITL \cite{Choi22} and WireSegHR \cite{Chiu23}. As previously discussed in \cref{sec:introduction}, these have their shortcomings, including: viewing perspective and setting different from the one encountered in manned aircraft flights \cite{Candamo09, Chiu23}, small size of the dataset with crops of larger images \cite{Choi22}, containing scaled down larger images, with low resolution and image-level labels \cite{Yetgin18a, Yetgin18b} or featuring clearly visible power lines \cite{Yetgin19, Zhang19h, Abdelfattah21}, oftenly seen from a short distance. The latter are more suitable to test algorithms designed to guarantee the safety of UAVs operating in the proximity of the power line infrastructure, mostly for inspection purposes. However, in this work, we want to perform the detection at longer ranges for collision-avoidance purposes. Furthermore, we are doing it in images captured by aircraft-mounted cameras with short focal lengths to cover the large area in front of the aircraft required to ensure safety.


\begin{figure*}[ht]
	\centering
    \includegraphics[width=\linewidth]{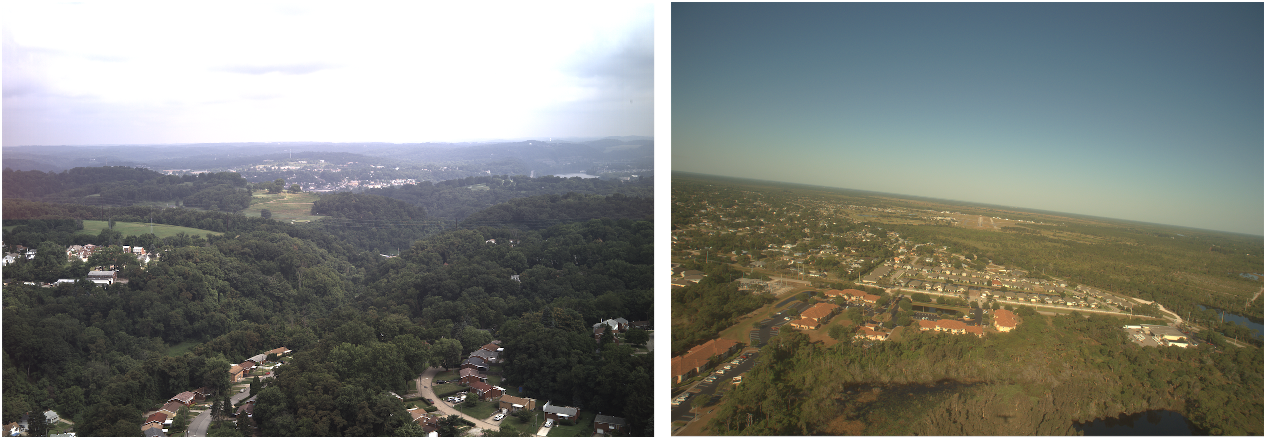}  
	\caption{Examples of the images from the \TheirDataset{} dataset~\cite{Stambler19} (left) and from our \OurDataset{} dataset (right).}
	\label{fig:sample}
\end{figure*}

To start developing an aircraft safety system for cable and pylon obstacles avoidance, we created and labeled our own dataset, which we will refer to as the \OurDataset{} dataset. The \TheirDataset{} dataset~\cite{Stambler19} is the publicly available dataset that most resembles ours and was created with the same purpose in mind. We tested our algorithm on both datasets and we provide additional details below.

\parag{\TheirDataset{} Dataset~\cite{Stambler19}.}
It is a public dataset containing 109 images of size up to $6576\times4384$ px. They were extracted from recordings captured by aircraft-mounted, forward facing cameras with a large focal length. The high resolution and the large focal length of the cameras allowed for the capture of power line cables at long ranges within a narrow field of view -- 25\textdegree{} horizontally and 18\textdegree{} vertically. The images were labeled using pixel-wise masks of power line cables. All data originates from flights conducted around Pittsburgh, PA in the USA.

\parag{\OurDataset{} Dataset.}
We built it ourselves from videos acquired during flights over Switzerland, the UK, Brazil, and the USA. It contains 865 images of size $4096\times3000$ px each. They were recorded by aircraft-mounted, forward facing cameras with a wide field of view -- up to 81\textdegree{} horizontally and 64\textdegree{} vertically. It is significantly wider than the one in the images from the \TheirDataset{} dataset. This makes the detection task more challenging but provides more context by covering a larger area in front of the aircraft. The videos were recorded at the rate of 6 frames per second, but the annotated frames were extracted with sufficient time intervals between them to provide for a diversified scene content. Polyline approximations of individual power line cables were generated by hand and bounding boxes were used to denote the pylons. Additionally, exclusion areas, that is, regions containing thin objects not associated with cables---flagpoles, cranes, communication towers---were marked by bounding boxes. We did not use them to create training data and excluded them from the pylon detection performance evaluation.

\subsection{Baselines}
\label{ssec:baselines}

We compare the performance of our models against three baselines: {\TheirModel}~\cite{Stambler19}, PIDNet-S~\cite{Xu23a} and {\RTDETR}~\cite{Zhao24}. We chose the first as the only method which has been so far evaluated for the detection of far-away cables on the \TheirDataset{} dataset. We chose the other two as modern representatives of semantic segmentation and object detection network architectures, respectively, and because they were designed with issues of inference speed in mind. This is an important consideration for the deployment of deep networks that must run in real-time on hardware that can be installed in an aircraft.

We compare cable detection performance obtained by our models against that of \TheirModel{} on both datasets, while the remaining comparisons are done on the \OurDataset{} dataset alone. 

We present each of the baselines in more detail below.

\parag{\TheirModel{} \cite{Stambler19} -- cable detection.}
It is a cable detection model introduced jointly with the \TheirDataset{} dataset. We reimplemented the CNN as described in the publication. However, we did not reimplement the post-processing steps involving clustering of the regressed line segments and a refinement with moving least squares which ultimately led to the generation of segmentation masks at input image resolution. This makes it simple to compare the performance of \TheirModel{} against ours using the coarse resolution predictions of both models. They are 16 times spatially downsampled with respect to the input image size.

\parag{PIDNet-S~\cite{Xu23a} -- cable detection.}
This is a member of a family of models designed for semantic segmentation. We train it without boundary-aware loss components, thus also without D branch supervision. We did this because these components are mostly useful for multi-class semantic segmentation with multiple, potentially confusing, object boundaries in the image. These issues do not arise for power line cables. We start from the ImageNet-pretrained checkpoint of the PIDNet-S model and add at the end of it a single $2\times2$ strided convolution layer to match the downsampling factor of our models. We also replaced the SGD optimizer with Adam \cite{Kingma14} and did not use the poly learning rate strategy.

\parag{\RTDETR{} \cite{Zhao24} -- pylon detection.}
This object detector uses the ResNet18 backbone~\cite{He16a}. As PIDNet-S, it has been designed with real-time applications in mind. It builds on top of transformer-based object detectors. We additionally combine it with SAHI \cite{Akyon22}, training and inference strategy for object detection in high-resolution images. We evaluate two variants of patched input scales. In both cases, we run the object detector over full-resolution image and its split into $1024\times1024$ overlapping patches, appropriately resizing the inputs for the detector and combining all of its outputs. In the second variant we additionally use similarly generated sets of $512\times512$ and $256\times256$ patches and add their detection results to the output. For evaluation, we rescale the final predicted bounding boxes to the downsampled output resolution of our model, use them to generate binary segmentation masks and convert them into distance masks. Also, for this model we initialize it with the pretrained weights provided with the official model implementation, obtained from training on the Objects365~\cite{Shao19} dataset and subsequent fine-tuning on the COCO~\cite{Lin14a} dataset.

\parag{PIDNet-S \cite{Xu23a} -- joint cables and pylons detection.}
We added additional segmentation heads to three branches of the PIDNet-S model to enable joint prediction of cables and pylons segmentation masks, as in our own model. In the loss function, we duplicate the cross-entropy term for the newly added pylon branch and add a weighting hyper parameter by which the added pylon segmentation loss components are multiplied. As before, we do not use boundary-aware loss components, start from the pre-trained model and use the Adam optimizer without learning rate scheduler. To match the downsampling factor of our models the final segmentation outputs for cables and pylons are bilinearly downsampled before being passed to the loss function.

\subsection{Evaluation Metrics}
\label{sec:metrics}

To compare the different models, we computed pixel-wise precision, recall and {\FOne} score for the generated cable and pylon segmentation masks. To obtain these masks for our models from the predicted distance masks we simply threshold them at a selected distance value. 

We also report correctness, completeness and quality (CCQ) metrics \cite{Wiedemann98}. These metrics were originally applied to road extraction task and defined in terms of \emph{lengths} of the extracted and ground-truth road segments. We implement them with respect to the \emph{areas} of foreground objects, measured in the output resolution on the binary segmentations obtained from thresholded distance masks. Correctness and completeness operate as standard precision and recall, respectively, but include a controllable tolerance for small detection errors within the defined error margin. Those inaccuracies can stem from, for instance, imprecisions in localizing the object by the detection network or the $\LMalis$ loss component in cable detections, which can cause them by forcing mergers in predictions of disconnected cable line segments. In our experiments we set the tolerated error distance to a value corresponding to the closest 8-pixel neighborhood of the ground-truth segmentation foreground. In other words, the correctness and completeness should be considered as relaxed precision and recall, respectively. This relaxation can be implemented with dilation by 1 pixel of ground truth area for considering true positive detections, decreasing accordingly the area in which the detections are treated as false positives and counting as a false negative any exact ground-truth pixel which is not covered by 1-pixel-dilated exact prediction. Effectively, these metrics are insensitive to the spurious or lacking detections in the predicted segmentation masks if those are found within the tolerated error distance from the exact ground-truth.

We did not use a held out test set for either of the datasets. Instead, in the development of our model we mostly relied on a fixed \mbox{train/validation} split of the \OurDataset{} dataset, as well as on a 5-fold cross validation split. For the performance comparisons presented below, we report the validation performance obtained from running a single 5-fold cross validation for each model. Since both datasets contain temporally dependent image sequences, we partitioned the data into folds by splitting on the recordings level. Additionally, for the 5-fold cross validation we grouped the recordings so as to minimize the overlap of the same geographical locations within single folds. 

The reported metrics are for the model states at epochs at which the mean validation quality of the target object segmentation was the highest during each respective 5-fold cross validation run. The target objects are cables or pylons, depending on the specific performance comparison presented in the tables below. For models performing joint cables and pylons segmentation, the cables are selected as target objects and pylon detection metrics are reported as achieved at the respective epochs. We report the means and standard deviations of each metric computed across all validation folds.

\subsection{Performance Evaluation}
\label{ssec:performanceEvaluation}


\begin{table*}[t]
    \sisetup{
        table-alignment-mode = format, 
        table-number-alignment = center, 
        propagate-math-font = true,
        reset-math-version = false, 
        separate-uncertainty
    }
    \caption{Comparison of power line cables detection performance between our method and \TheirModel{} (\TheirModelAbbr) \cite{Stambler19} on the \TheirDataset{} and \OurDataset{} datasets. Results for the two size variants of our models are shown -- \SmallDDLN{} and \LargeDDLN. \SmallDDLN{} has 2.7 million trainable parameters, about the same amount as the \TheirModel{} model.}
    \label{tab:results}
    {\footnotesize
        \begin{tabularx}{\textwidth}{@{\extracolsep{\fill}} @{} *{2}{l} *{7}{S} @{}}
        \toprule
        Method & Object & {FPS} & {Precision} & {Recall} & {\FOne} & {Correctness} & {Completeness} & {Quality} \\
        \midrule

        & & & \multicolumn{6}{c}{{\TheirDataset{}}} \\
        \cmidrule{4-9}

        \TheirModelAbbr{} & cables & 9.2 & \num{0.71(7)} & \num{0.63(6)} & \num{0.66(6)} & \num{0.73(7)} & \num{0.72(7)} & \num{0.57(8)} \\
        \SmallDDLN{} & cables & 12.8 & {\boldmath \num{0.74(9)}} & \num{0.70(8)} & \num{0.71(4)} & {\boldmath \num{0.79(6)}} & \num{0.79(7)} & \num{0.65(5)} \\
        \LargeDDLN{} & cables & 9.1 & \num{0.73(7)} & {\boldmath \num{0.83(6)}} & {\boldmath \num{0.78(6)}} & {\boldmath \num{0.79(7)}} & {\boldmath \num{0.90(5)}} & {\boldmath \num{0.72(7)}} \\
        
        \cmidrule{4-9}
        & & & \multicolumn{6}{c}{{{\OurDataset}}} \\
        \cmidrule{4-9}

        \TheirModelAbbr{} & cables & 9.2 & \num{0.52(9)} & \num{0.42(6)} & \num{0.47(7)} & \num{0.54(9)} & \num{0.49(6)} & \num{0.35(7)} \\
        \SmallDDLN{} & cables & 12.8 & \num{0.56(12)} & \num{0.64(4)} & \num{0.59(9)} & \num{0.65(14)} & \num{0.74(3)} & \num{0.53(10)} \\
        \LargeDDLN{} & cables & 9.1 & {\boldmath \num{0.71(6)}} & \num{0.73(5)} & \num{0.72(5)} & {\boldmath \num{0.83(5)}} & \num{0.80(4)} & \num{0.69(6)} \\

        \addlinespace[0.5em]

        \SmallDDLN{} & cables & {\multirow{2}*{12.8}} & \num{0.59(8)} & \num{0.60(5)} & \num{0.59(6)} & \num{0.67(8)} & \num{0.70(5)} & \num{0.52(7)} \\
        (joint) & pylons &  & {\boldmath \num{0.41(8)}} & \num{0.40(7)} & \num{0.39(4)} & \num{0.68(7)} & \num{0.56(10)} & \num{0.43(6)} \\  
        
        \addlinespace[0.5em]

        \LargeDDLN{} & cables & {\multirow{2}*{9.1}} & {\boldmath \num{0.71(5)}} & {\boldmath \num{0.75(5)}} & {\boldmath \num{0.73(4)}} & {\boldmath \num{0.83(4)}} & {\boldmath \num{0.82(4)}} & {\boldmath \num{0.70(4)}} \\
        (joint) & pylons &  & {\boldmath \num{0.41(10)}} & {\boldmath \num{0.61(4)}} & {\boldmath \num{0.48(8)}} & {\boldmath \num{0.71(11)}} & {\boldmath \num{0.76(3)}} & {\boldmath \num{0.58(7)}} \\  

        \bottomrule
        \end{tabularx}
    }
    \begin{minipage}{\textwidth}
        \vspace{0.25em}
        {\footnotesize \textbf{Bolded results} highlight the best mean metric value within each group of the given dataset and the detected object type.}
    \end{minipage}
\end{table*}


\begin{figure*}[!ht]
	\centering
	\includegraphics[width=0.8\linewidth]{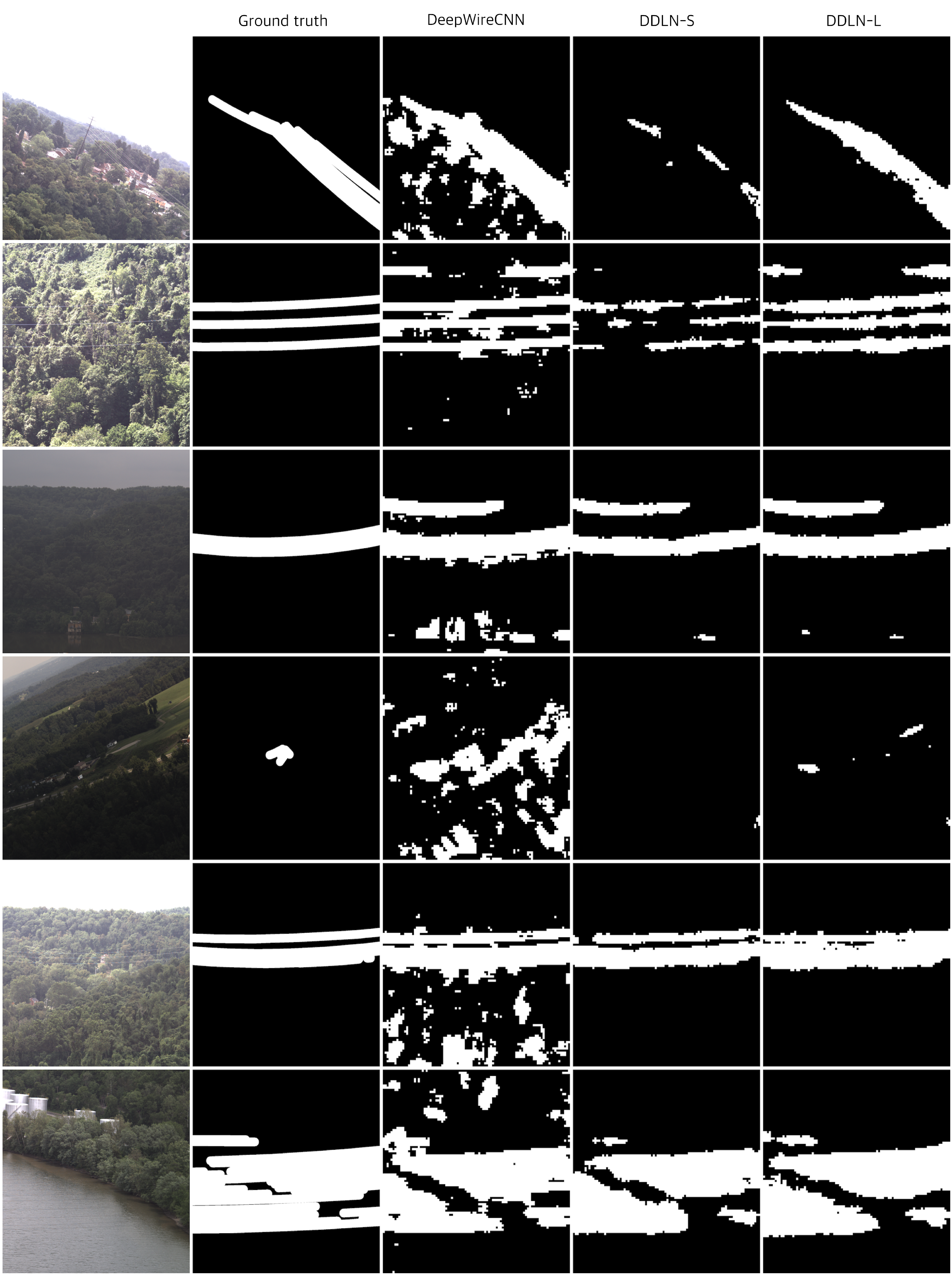}
	\caption{Exemplary predictions of the three models trained on the \TheirDataset{} dataset~\cite{Stambler19}: \TheirModel{}~\cite{Stambler19}, \SmallDDLN{} and \LargeDDLN. The inputs are $1536\times1536$ px crops from full-resolution images from the validation set. Predictions were generated by models initialized from checkpoints at epochs with highest validation cable segmentation quality metric.}
	\label{fig:nevbwPredictions}
\end{figure*}

\parag{Cable Only Detection.} 

\Cref{tab:results} shows the main performance comparison between the two variants of our model and the \TheirModel{}~\cite{Stambler19}, both on the \TheirDataset{} dataset and our \OurDataset{} dataset. We also measured inference speed as a number of processed $1\times3\times4096\times3000$ random input tensors per second (i.e. frames per second, FPS) on a NVidia V100 GPU, using automatic mixed-precision arithmetic. Additionally, \Cref{fig:nevbwPredictions} depicts qualitative results. After tuning the hyperparameters of the \TheirModel{} model we achieved 0.70$\pm$0.09 AP for the segmentation of cables at coarse resolution and the values for the remaining metrics as reported in the first row of the \Cref{tab:results}. The results on \TheirDataset{} dataset demonstrate that both the small and the large variant of our model outperform \TheirModel{} on all cable detection metrics. Additionally, they do so without increasing the inference time. We also observed that our models achieve high performance scores earlier in the learning process than the \TheirModel{}. We attribute this to the initialization of the networks with the pretrained weights. Focusing only on our models, it is also clear that the larger variant achieves better or equivalent performance to its smaller counterpart across all cable detection metrics.


\begin{figure*}[!ht]
	\centering
	\includegraphics[width=\linewidth]{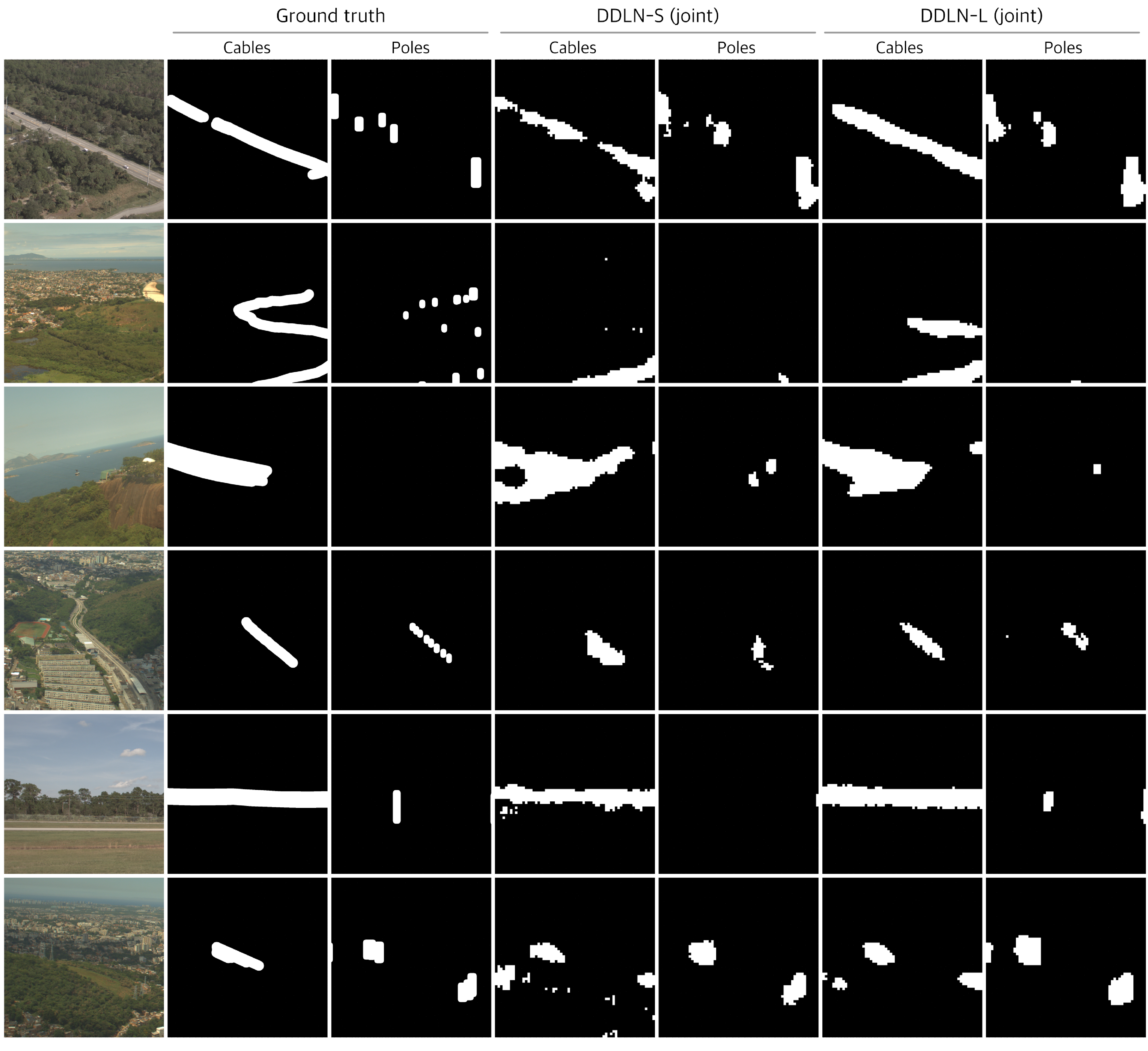}
	\caption{Exemplary predictions of the two size variants of our model trained jointly for cable and pylon detection on the \OurDataset{} dataset. The inputs are $1024\times1024$ px crops from full-resolution images from the validation set. Predictions were generated by models initialized from checkpoints at epochs with highest validation cable segmentation quality metric.}
	\label{fig:ddlnJointPredictions}
\end{figure*}

\begin{table*}[t]
    \sisetup{
        table-alignment-mode = format,
        table-number-alignment = center,
        propagate-math-font = true,
        reset-math-version = false,
        separate-uncertainty
    }

    \caption{Comparison of cable detection performance on \OurDataset{} dataset between our models and PIDNet-S \cite{Xu23a}, a modern semantic segmentation network for real-time applications. PIDNet-S was trained for cable detection only, while our models were trained jointly for cables and poles detection.}
    \label{tab:baselineCables}

    {\small
        \begin{tabularx}{\textwidth}{@{\extracolsep{\fill}} @{} *{1}{l} *{8}{S} @{}}
        \toprule
        Method & {Size} & {FPS} & {Precision} & {Recall} & {\FOne} & {Corr.} & {Compl.} & {Quality} \\
        \midrule
        
        PIDNet-S & 7.6 & 22.2 & \num{0.74(7)} & \num{0.69(5)} & \num{0.71(6)} & \num{0.81(7)} & \num{0.76(4)} & \num{0.65(7)} \\
        \MediumDDLN{} & 7.5 & 10.7 & \num{0.71(7)} & \num{0.71(5)} & \num{0.71(6)} & \num{0.83(6)} & \num{0.80(4)} & \num{0.68(6)} \\

        \bottomrule
        \end{tabularx}
    }
    
    \vspace{-0.25em}
\end{table*}

In \Cref{tab:baselineCables} we compare the cable detection performance of our \MediumDDLN{} model against that of PIDNet-S~\cite{Xu23a}. As the latter is designed for real-time performance, we also report the model size and inference speeds for the networks. The size is given as the number of millions of trainable parameters in the respective model. Regarding detection performance, our \MediumDDLN{} model, which has the same size as PIDNet-S, yields slightly better cable detection quality and a similar  \FOne{} score. Given this result, we conducted additional experiments with PIDNet-S trained to simultaneously detect cables and pylons, as previously described in \cref{ssec:baselines}, and compared its performance on both tasks with our joint detection model, which we detail below. 


\begin{table*}[t]
    \sisetup{
        table-alignment-mode = format,
        table-number-alignment = center,
        propagate-math-font = true,
        reset-math-version = false,
        separate-uncertainty
    }

    \caption{Comparison of joint cables and pylons detection performance on \OurDataset{} dataset between \MediumDDLN{} and PIDNet-S \cite{Xu23a}. PIDNet-S was trained for cable and pylon detection jointly as described in \cref{ssec:baselines}. The reported metrics are measured at the epochs with the highest validation cable detection quality.}
    \label{tab:baselineCablesAndPoles}

    {\small
        \begin{tabularx}{\textwidth}{@{\extracolsep{\fill}} @{} *{2}{l} *{6}{S} @{}}
        \toprule
        Method & Object & {Precision} & {Recall} & {\FOne} & {Corr.} & {Compl.} & {Quality} \\
        \midrule
        
        \multirow{2}*{PIDNet-S} & cables & \num{0.70(5)} & \num{0.68(4)} & \num{0.69(4)} & \num{0.81(3)} & \num{0.76(4)} & \num{0.65(5)} \\
        & pylons & {\boldmath \num{0.55(4)}} & {\boldmath \num{0.59(2)}} & {\boldmath \num{0.57(3)}} & \num{0.71(4)} & {\boldmath \num{0.68(2)}} & \num{0.53(3)} \\
        \multirow{2}*{\MediumDDLN} & cables & {\boldmath \num{0.73(6)}} & {\boldmath \num{0.70(6)}} & {\boldmath \num{0.71(5)}} & {\boldmath \num{0.86(6)}} & {\boldmath \num{0.78(5)}} & {\boldmath \num{0.69(7)}} \\
        & pylons & \num{0.52(11)} & \num{0.56(4)} & \num{0.53(6)} & {\boldmath \num{0.74(10)}} & \num{0.66(4)} & {\boldmath \num{0.54(5)}} \\

        \bottomrule
        \end{tabularx}
    }
\end{table*}

\parag{Joint Cable and Pylon Detection.} 

\begin{figure*}[ht]
	\centering
	\includegraphics[width=\linewidth]{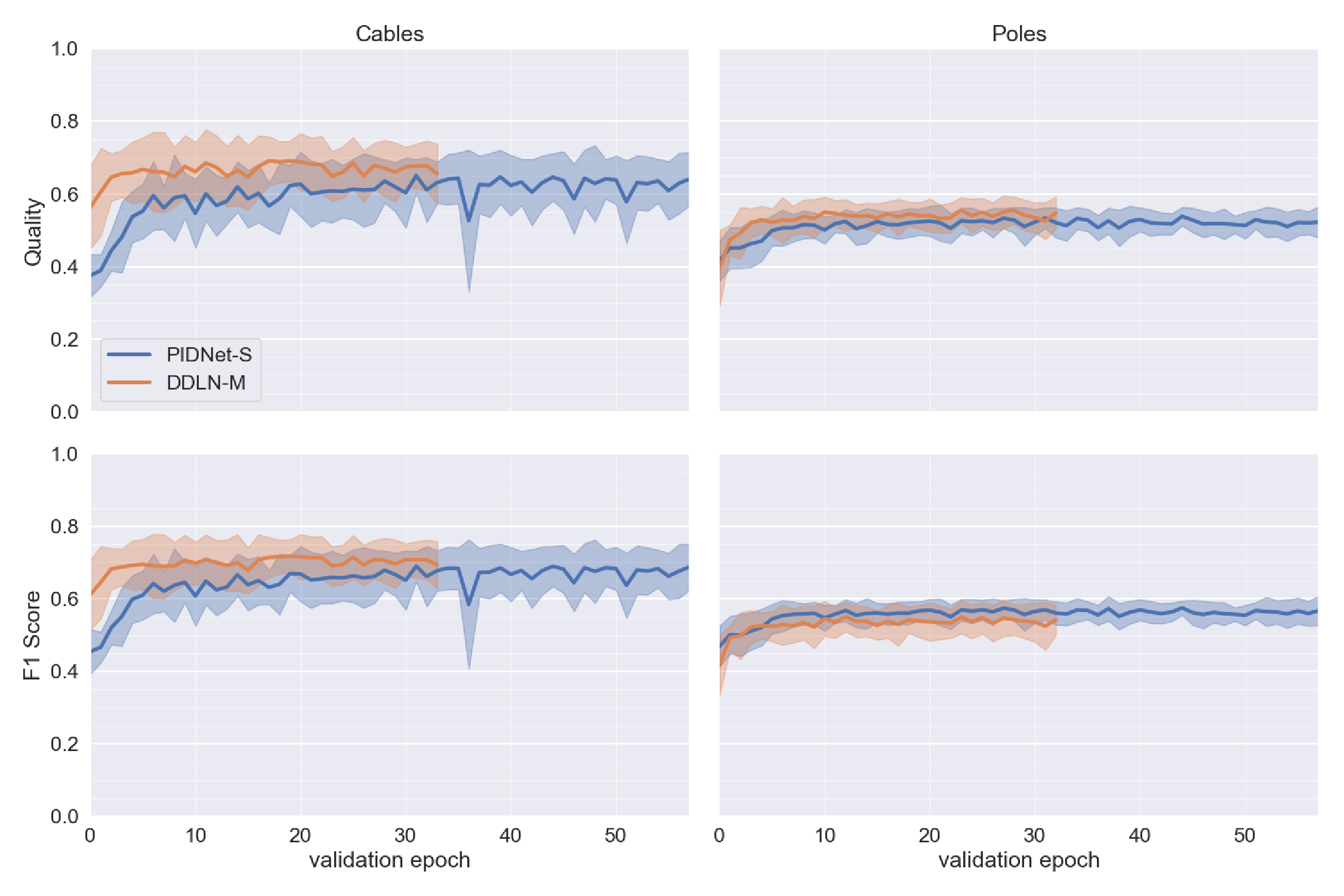}
	\caption{Evolution of validation metrics -- quality and {\FOne} for cable and pylon detection -- in 5-fold cross validation runs reported in \cref{tab:baselineCablesAndPoles} for \MediumDDLN{} and PIDNet-S~\cite{Xu23a}, both trained for joint detection task. The solid lines are means of respective metrics, while the shaded areas correspond to standard deviations.
    \label{fig:plot}}
\end{figure*}

To continue the  comparison of our method with the PIDNet-S~\cite{Xu23a} baseline, we added to the latter a second segmentation output head to enable it to also perform pylon detection. We report our results in \Cref{tab:baselineCablesAndPoles} and shown in \Cref{fig:plot} the evolution of the validation score during training. Our {\MediumDDLN} model still outperforms PIDNet-S on the cable detection quality, as well as on the previously equal {\FOne} score. It also delivers minimally higher pylon detection quality, while it records a slightly lower {\FOne} score than PIDNet-S. Identical observations can be made by looking at the \Cref{fig:plot} where validation performance curves of our {\MediumDDLN} model are consistently, and across all validated epochs, above those of PIDNet-S for cable detection quality and {\FOne} score.  As in the metrics reported in \Cref{tab:baselineCablesAndPoles}, pylon detection quality of {\MediumDDLN} is also minimally above that of PIDNet-S, while pylon detection {\FOne} score is slightly below the one of the segmentation baseline. Furthermore, recall from  \Cref{tab:results} that {\LargeDDLN} performs even better. 

Regarding the network inference speeds, PIDNet-S is the faster one with 22.2 FPS compared to our 10.7. However, this high speed provided by PIDNet-S may not be absolutely necessary for a possible deployment of our method. We have successfully integrated {\SmallDDLN}, the less performant but faster smaller alternative of our model, in the hardware system operating in real-time onboard of a helicopter, as described in \cref{ssec:hardware}. We run the system in flight and reached average inference speed of 5 FPS. This is already a good running speed for the application purpose of an aircraft safety system.

Additional results for our models trained jointly for cables and pylons detection on the \OurDataset{} dataset are shown at the bottom of \Cref{tab:results}. Qualitative results are shown in \Cref{fig:ddlnJointPredictions}. The cable detection performance metric is not significantly affected by adding joint detection of pylons. It can be hypothesized that the model trained for cable detection alone has already implicitly learned to rely on pylon presence contextual cues. Alternatively, the models trained for the joint prediction leverage the supervision for the pylon detection task to adapt their cable detection outputs, but affecting the detection of power lines both positively and negatively, with both contributions equalizing in the final scores. However, as for independent cable detection, pylon detection performance also improves with increased model size.

In other words, relying on pylon detections for cable detection, which the network may learn to do in joint prediction setting, can be a double-edged sword. Power line cable presence does not have to be accompanied by a visible presence of pylon structures. Multiple occlusion scenarios can occur when the pylons exist but are hidden behind terrain, buildings etc. or are outside of the current field of view. Furthermore, for cables close to the camera, the associated pylons, even if contained within the camera frustum, may fall out of the receptive field of the network for the given cable section. Moreover, the power line cables can also be attached to other structures, e.g. mounted on the rooftops of the buildings. Finally, the cables or other wiry objects which also pose the danger to the low-altitude aircraft do not have to be transmission lines \textit{per se}. As such they are not related to the presence of transmission towers. Therefore, in some scenarios pylon detection may be helpful in finding poorly visible cables by inferring their presence from the better visible pylons to which these cables are most likely attached. In other situations, the model could rely too heavily on pylon detections to determine whether the power line cable exists at a certain location and fail to detect it independently.


\parag{Pylon Only Detection.}

Finally, we compare our models against \RTDETR{}~\cite{Zhao24} in \cref{tab:baselinePoles} focusing on the pylon detection alone. In the presented results we adapt two kinds of thresholds used to turn \RTDETR{} predictions into distance masks: a cutoff confidence threshold filtering out the least certain bounding box predictions and a distance mask threshold used to convert a distance mask, derived from predicted bounding boxes, into a segmentation mask. The choice of these hyperparameters is different depending on whether we optimize for best quality or {\FOne} score, therefore we report both, underlining the metric with respect to which we optimized the thresholds. 

Firstly, looking at the two variants of inference done with the \RTDETR{} object detector we find that using either multiple scales of overlapping input image patches (with sizes of $256 \times 256$ px, $512 \times 512$ px and $1024 \times 1024$ px) or just a single scale ($1024 \times 1024$ px patches) does not change the obtained final performance much, with single scale providing with slightly higher performance. Using multiple scales affects greatly the inference speed of the model, as each level of smaller patch size adds roughly four times more input images to process per frame than found on the previous patch size level.

Comparing \RTDETR{} with our approach, we note that if we grow the size of our model to nearly match the size of this object detector, our network outperforms it on correctness, completeness, quality and recall metrics and falls short on segmentation precision and {\FOne} score. Focusing on the precision, we can observe though that its relaxed version---correctness---is much larger than its exact version for our models. That indicates that a large part of the pylon pixels predicted by our \HugeDDLN{} model and treated as false positives is actually away from the pylon segmentation ground-truth only by the distance smaller or equal to the error margin, as previously described in \cref{sec:metrics}. 

One reason for our superior performance on the CCQ metrics is due to the fact that those give, by design, more favourable evaluation to the method inherently using distance mask regression than to the one in which the distance masks are only derived from a different representation -- precisely regressed bounding boxes, as done for \RTDETR{}. The distance mask regression objective imposes a distance-dependent structure on the outputs regressed by our networks. This structure is also maintained when a segmentation mask is obtained through distance mask thresholding. Finally, the CCQ metrics, which relax standard precision and recall, also have a distance-dependent component in their use of error tolerance margin, which interplays favourably with the imposed distance-dependent structure of our targets. Meanwhile, the distance masks generated for \RTDETR{} are only derived with exact Euclidean distance transform from originally predicted bounding boxes. Therefore, they do not naturally inherit the distance-dependent structure in the resulting distance mask representation of the object detector's original output. 

We hypothesize that another reason for the favourable CCQ evaluation of the distance mask approach can be found in the cases concerning distant pylons clustered into groups. Those are not individually distiguishable on the predicted distance masks and an inaccurate or incomplete segmentation of partially detected pylon groups may not affect the metrics negatively. For example, if spurious foreground detections happen to fall into the region corresponding to the less clearly detected neighboring pylon, this will not have a negative impact on the CCQ metrics. 

Nonetheless, our model performs well in comparison to \RTDETR{}, while it is simultaneously successfully detecting cable presence, using the same distance regression based method. The joint detection of power line cables is something that an object detector like \RTDETR{} is not well suited to do. The cables are extremely elongated and thin objects, found in various orientations. Detecting them with an object detector method would mean that the ground truth bounding boxes would have highly variable size, be frequently largely overlapping with each other or would need to encompass cable groups as single objects and that they would contain mostly background objects and texture within their bounds.


\begin{table*}[t]
    \sisetup{
        table-alignment-mode = format,
        table-number-alignment = center, 
        propagate-math-font = true,
        reset-math-version = false, 
        separate-uncertainty
    }

    \caption{Comparison of pylon detection on \OurDataset{} dataset between our distance mask regression method and \RTDETR{}~\cite{Zhao24} combined with SAHI~\cite{Akyon22}. Our models were trained jointly for cables and pylons detection. The metrics for our models are reported for epochs with the best mean validation pylon detection quality. For \RTDETR{} runs, the reported epochs were selected as ones with maximal mean validation value of the \underline{underlined} metric -- quality or \FOne. The outputs of the models are evaluated with distance masks of 32 times smaller resolution than the input image size, corresponding to the size of the output of the \HugeDDLN{} model.}
    \label{tab:baselinePoles}

    {\footnotesize  
        \begin{tabularx}{\textwidth}{@{\extracolsep{\fill}} @{} *{1}{l} *{8}{S} @{}}
        \toprule
        Method & {Size} & {FPS} & {Precision} & {Recall} & {\FOne} & {Corr.} & {Compl.} & {Quality} \\
        \midrule
        {\multirow{2}*{\RTDETR}} & {\multirow{2}*{\tablenum{20.1}}} & {\multirow{2}*{\tablenum{0.9}}} & \num{0.62(12)} & \num{0.50(9)} & \underline{\num{0.55(9)}} & \num{0.71(14)} & \num{0.58(12)} & \num{0.47(12)}\\
         &  &  & \num{0.25(5)} & \num{0.50(9)} & \num{0.33(7)} & \num{0.74(10)} & \num{0.75(5)} & \underline{\num{0.60(5)}} \\
        {\multirow{2}*{\RTDETR\textsuperscript{s}}} & {\multirow{2}*{\tablenum{20.1}}} & {\multirow{2}*{\tablenum{16.5}}} & {\boldmath \num{0.64(6)}} & \num{0.50(9)} & \underline{\boldmath \num{0.56(8)}} & \num{0.74(9)} & \num{0.58(6)} & \num{0.49(6)} \\
         &  &  & \num{0.26(5)} & \num{0.52(8)} & \num{0.34(6)} & \num{0.75(7)} & \num{0.76(5)} & \underline{\num{0.61(5)}} \\
        \HugeDDLN & 18.3 & 8.8 & \num{0.45(7)} & {\boldmath \num{0.61(3)}} & \num{0.51(5)} & {\boldmath \num{0.83(4)}} & {\boldmath \num{0.79(4)}} & {\boldmath \num{0.68(3)}} \\

        \bottomrule
        \end{tabularx}
    }
    
    \vspace{-0.5em}
    \begin{minipage}[t]{\textwidth}
        \begin{flushleft}
        {\footnotesize
            \textsuperscript{s} -- single scale of input image patches (1024 $\times$ 1024). 
        }
        \end{flushleft}
    \end{minipage}
\end{table*}

\subsection{Implementation Details}
\label{ssec:implementation}

\parag{Distance and segmentation masks.}
To generate learning targets for our models, we converted the annotations into distance masks by computing unsigned Euclidean distance transform to the detected objects (cables and pylons). For the \OurDataset{}, we treat as object areas 5-pixel-thick regions around annotation polylines and the insides of the pylons' bounding boxes. The target distance regression masks are first computed in pixel distance units, then clamped to the maximum distance value of 128 and finally normalized to the $[0, 1]$ range. The distance masks are downsampled through $16\times16$ min-pooling operation to provide the targets for learning at the coarse resolution. 

\noindent To generate segmentation masks used by our baselines -- \TheirModel{}~\cite{Stambler19} and PIDNet-S~\cite{Xu23a}, we binarized full resolution distance masks with a distance value threshold of 32. Then we downsampled the labels to match the output resolution. To perform the $n$ times downsampling of the segmentation masks we look at the center 4 pixels of $n \times n$ patches and if they contain a foreground object we label the pixel in the downsampled output as foreground as well, and as background otherwise.

\parag{Patch sampling.}
For the experiments run on \TheirDataset{} dataset and also for those using \TheirModel{}, we followed the patch sampling procedure proposed by \citet{Stambler19}. For all other experiments, we use our own sampling method. We select patch center locations dynamically during training from the high-resolution images, based on the data annotations. Patches are sampled as centered on some part of the target detected object or within its vicinity, with a set maximal allowed distance to the closest object and ensuring that the entire patch remains within the image boundaries. This effectively implements sampling position with random horizontal and vertical jitter. When training our cable detection and joint cable and pylon detection models we do not sample purely background patches. We did so only when training the baselines models, where this setting was one of the hyperparameters. In most experiments we centered the sampled patches on power line cables. This procedure was different when training pylon-only detection model -- {\RTDETR}~\cite{Zhao24} -- where we analogously centered the sampled patches on pylons instead of cables. It was also different when training PIDNet-S \cite{Xu23a} and DDLN-M models for comparing the joint detection performance in \cref{tab:baselineCablesAndPoles}. In those experiments we centered patches within the union of close-to-cables and close-to-pylons areas.

\parag{Data augmentations.}
For the models trained on the \TheirDataset{} dataset, we used the set of augmentations proposed in \cite{Stambler19}. For all our models trained on the \OurDataset{} dataset, as well as for PIDNet-S~\cite{Xu23a}, we use color jittering and random horizontal flipping. Since the weights of these models were initialized with those obtained by pretraining on the ImageNet~\cite{Deng09} dataset, we normalize the input with the standard ImageNet mean and variance statistics. The \RTDETR{} was trained with a larger set of augmentations, including color jittering, random zoom-out, IoU crop and horizontal flipping; we refer the interested reader to the work of \citet{Zhao24} and the official implementation of this object detector where this data augmentations set can be found. All the models were trained with large image patches of $1024\times1024$ px size.

\parag{Hyperparameters.}
For all models we compare against, we tuned selected sets of their hyperparameters on one of the splits of the \TheirDataset{} dataset or on our \mbox{train/validation} development split of \OurDataset{}. For this procedure we used SMAC3~\cite{Lindauer22}, a framework implementing Bayesian optimization and Hyperband~(BOHB)~\cite{Falkner18} for hyperparameter optimization. To obtain the reported performance metrics values we run the 5-fold cross validation with the model and data augmentations configured according to the found best performing hyperparameters configurations. Finally, when thresholding on the probabilities predicted by \TheirModel{}~\cite{Stambler19} and PIDNet-S~\cite{Xu23a} models, as well as on confidence scores and on distance masks derived from \RTDETR{}~\cite{Zhao24} predictions, we also choose threshold values which maximize the validation metrics of the obtained target object segmentation. We choose these threshold values for each data fold separately, maximizing their respective validation performance, therefore the reported mean metrics represent the best achievable performance, providing very strong baseline results for our models to compare with.

For our cable detection models, we selected the hyper parameters firstly mostly manually, throughout the method development process, and then by trying out several configurations and running with them the complete 5-fold cross validation experiments. Finally, we selected the configurations achieving the best maximal mean validation cable or pylon segmentation quality. We also selected thresholds, same for all folds in a single 5-fold cross validation run, for converting predicted distance masks into segmentations so as to maximize the detection validation quality metric, both for cables and pylons. We resorted to some automated tuning with SMAC3~\cite{Lindauer22} for our comparison with PIDNet-S~\cite{Xu23a} for joint cables and pylons detection.

We note that selection of these hyperparameters is important for the final performance. In the process we found that with improperly selected values, the \SmallDDLN{} model achieved on the \TheirDataset{} dataset the performance equivalent to that of \TheirModel{} model. 

\subsection{Ablations}

To demonstrate the impact of the components used in the proposed method on the performance of power line cable detection, we run ablation experiments with the \SmallDDLN{} model trained on the \TheirDataset{} dataset for independent cable detection. Their results are presented in the \cref{tab:ablation}.


\begin{table*}[t]
    \sisetup{
        table-alignment-mode = format,
        table-number-alignment = center,
        propagate-math-font = true,
        reset-math-version = false,
        separate-uncertainty
    }

    \caption{Ablation study demonstrating the impact of the components in our proposed method on cable detection performance. The ablation experiments have been conducted with the \SmallDDLN{} model on the \TheirDataset{} \cite{Stambler19} dataset. ``Pretr.'' indicates the initializing the model parameters with the weights of \ConvNextTiny{} pretrained on the ImageNet \cite{Deng09} dataset. ``LIF'' refers to logarithmically scaled inverse frequency weights in $\LDat$. $\LMalis$ \cite{Oner22a} referes to the loss term promoting connectivity in cable detections.}
    \label{tab:ablation}
    
    {\small
        \begin{tabularx}{\textwidth}{@{\extracolsep{\fill}} @{} *{3}{m{0.05\textwidth}<\centering} *{6}{S} @{}}
        \toprule
        {Pretr.} & {LIF} & {$\LMalis$} & {Precision} & {Recall} & {\FOne} & {Correctness} & {Completeness} & {Quality} \\
        \midrule
        \checkmark & $-$ & $-$ & \num{0.85(8)} & \num{0.49(17)} & \num{0.60(14)} & \num{0.88(6)} & \num{0.56(19)} & \num{0.51(16)} \\
        \checkmark & \checkmark & $-$ & \num{0.75(11)} & \num{0.67(8)} & \num{0.70(4)} & \num{0.78(9)} & \num{0.75(9)} & \num{0.61(5)} \\
        \checkmark & $-$ & \checkmark & \num{0.67(15)} & \num{0.71(9)} & \num{0.67(7)} & \num{0.74(13)} & \num{0.81(9)} & \num{0.62(8)} \\
        \checkmark & \checkmark & \checkmark & \num{0.74(9)} & \num{0.70(8)} & \num{0.71(4)} & \num{0.79(6)} & \num{0.79(7)} & \num{0.65(5)} \\
        \bottomrule   
        \end{tabularx}
    }
\end{table*}

Initializing the model with pretrained weights and training it with simple MSD loss provides a highly precise model -- with high precision and correctness scores. However, the same model has low recall and completeness, missing a lot of cable detections, which in turn negatively affects the quality metric. We note however, that the initialization of our \ConvNextTiny{} backbone with pre-trained weights is indispensable to allow for successful training of our model -- we did not succeed in training this model in the same manner but from scratch.

The results are similarly improved by either adding the inverse frequency based weight terms, as in $\LDat$ from \cref{eq:LDat}, or by adding just $\LMalis$, the loss term promoting cable detections connectivity, to the plain MSD loss. Thanks to these additions the recall and the completeness increase significantly compared to the model trained only with the simple MSD loss.

For our complete setup we used both the $\LDat$ and $\LMalis$, which is the configuration for the model presented in all result tables apart from \cref{tab:ablation}. However, comparing the evolution of the validation cable segmentation quality metric during training for the complete setup and for those using either of the loss function components in isolation, it is apparent that these configurations are actually leading to equal performance for cable-only detection. We also note that the final scores depend here on the value of the used binarization threshold with which we convert distance masks into segmentation masks. In our experiments we observed that with a lower threshold value, using solely $\LMalis$ component gives better cable segmentation quality than $\LDat$ alone or both loss components jointly. However, overall the performance was lower than in the ablation experiments presented here in which we used a larger binarization threshold. Ultimately, for all the runs of our models reported in the previously described result tables we used both loss components and the larger binarization threshold value.


\section{Conclusion}
\label{sec:conclusion}

We have proposed a vision-based approach to joint detection of power line cables and transmission towers as seen by low-flying aircraft. To this end, we have combined a modern convolutional architecture with transfer learning and a loss function adapted to curvilinear structure delineation~\cite{Oner22a}. We have used a single network for both detection tasks and have demonstrated its performance on two benchmarking datasets.

As all methods, ours suffer from some limitations, some inherent to the general vision-based detection methodology and others that can be remedied by further work. Firstly, our method yields localized detections of power line cables and pylons in the form of segmentation masks. For training, we rely on bounding box annotations of the pylons, which are not particularly accurate representations for the not distant ones. More precise segmentation masks could be used instead in situations where the pylons' structure and background can be easily separated. Another avenue worth pursuing is to study how much the model relies on pylon predictions for cable detections, and {\it vice versa}, as  discussed in \cref{ssec:performanceEvaluation}. This aspect is important when deploying our approach in unusual scenarios that may not exhibit the same correlations as our training set. Also vertical obstacles that can be confused for pylons should be studied separately. Finally, the appearance of power line cables and pylons naturally varies under weather and lighting conditions, and also between different geographical locations with differing power lines infrastructure. Any vision-based deep learning method, such as ours, relies on a model trained on an ultimately limited dataset and therefore with a limited set of appearances for the obstacles that it is capable of handling. 

Increasing the inference speed of our model should also constitute a direction for future work. To achieve it, modifications in architectural components of the model backbone should be considered, a single example of which can be found in InceptionNeXt~\cite{Yu24b} regarding depth-wise convolutions. Incorporating attentional components into the model architecture, as can be found in PIDNet~\cite{Xu23a}, also presents a viable research direction, with the aim of increasing the receptive field and the expressivity of the network.

	\backmatter

\section*{Declarations}

\subsection*{Funding}
This work was supported under an Innosuisse Grant number 57140.1 IP-ICT funding the collaboration between Daedalean AG and EPFL.

\subsection*{Competing Interests}
\parag{Employment and financial interests} Ad Eberhard, Ivan Egorov, Philipp Kr{\"u}si and Grigory Yakushev declare that this work was done as part of their employment with Daedalean AG. Jakub Gwizda\l{}a, Doruk Oner, Soumava Kumar Roy, Mian Akbar Shah and Pascal Fua declare they have no financial interests.

\parag{Non-financial interests} None.

\subsection*{Dual-use}
The authors declare that the solutions and results of this work are intended for and were developed for commercial/industrial civilian aircraft operations. They shall not be used in any context in which any harm done to human lives, property and/or infrustructure as well as to natural habitats and ecosystems is a direct or an indirect consequence of their use.

\subsection*{Data}
This work used publicly available Near Earth Autonomy Vision Based Wire Detection (NE-VBW) dataset which is described at \url{https://sites.google.com/nearearth.aero/ne-vbwd} and can be downloaded from \url{https://drive.google.com/drive/folders/1A4hViJWaf9mwSISBAG629CGg_VnVHRtb}. Used 5-fold cross validation split of this dataset content can be accessed at \url{https://drive.google.com/file/d/1eodtX6bU-6lBSz1-rlIL5ptocu8kqV3t/view?usp=sharing}. The collected DDLN dataset remains proprietary to Daedalean AG and is not publicly available.

\subsection*{Authors' Contribution Statements}
Conceptualization: Grigory Yakushev, Pascal Fua; Funding acquisition: Pascal Fua; Data preparation: Ad Eberhard, Grigory Yakushev; Methodology: Doruk Oner, Pascal Fua, Grigory Yakushev; Investigation: Jakub Gwizda\l{}a, Mian Akbar Shah, Soumava Kumar Roy; Project administration: Pascal Fua, Ad Eberhard, Grigory Yakushev; Software: Jakub Gwizda\l{}a, Ivan Egorov, Mian Akbar Shah, Doruk Oner; Supervision: Grigory Yakushev, Pascal Fua, Doruk Oner; Writing -- original drafts: Jakub Gwizda\l{}a, Ivan Egorov, Pascal Fua; Writing -- review \& editing: Pascal Fua, Doruk Oner, Ivan Egorov, Philipp Kr{\"u}si, Soumava Kumar Roy.

\end{document}